# Continual Uncertainty Learning for Robust Control of Nonlinear Systems with Multiple Heterogeneous Uncertainties

Heisei Yonezawa[#1], Ansei Yonezawa[#2], Itsuro Kajiwara [#3]

#1(Corresponding author)
Division of Mechanical and Aerospace Engineering, Hokkaido University
N13, W8, Kita-ku, Sapporo, Hokkaido 060-8628, Japan

#2
Faculty of Information Science and Technology, Hokkaido University
Kita 14, Nishi 9, Kita-ku, Sapporo, Hokkaido 060-0814, Japan

#3
Division of Mechanical and Aerospace Engineering, Hokkaido University
N13, W8, Kita-ku, Sapporo, Hokkaido 060-8628, Japan

**Abstract**

Robust control of mechanical systems with multiple uncertainties remains a fundamental challenge, particularly when nonlinear dynamics and operating-condition variations are intricately intertwined. Although deep reinforcement learning combined with domain randomization has shown promise in mitigating the sim-to-real gap, simultaneously handling all the sources of uncertainty often leads to sub-optimal policies and poor learning efficiency. This study proposes continual uncertainty learning (CUL), a curriculum-based continual learning framework for robust control of nonlinear systems on which multiple heterogeneous uncertainties are simultaneously superimposed. The core idea is to decompose the original control problem into a sequence of continual learning tasks by extending the system into a set of plants whose uncertainties are progressively expanded and diversified, so that the strategy for handling each uncertainty is acquired sequentially. Within this curriculum, the policy is updated across the plant sets under a memory-efficient anti-forgetting regularization, which preserves the strategies acquired for earlier uncertainties. In parallel, a model-based controller that guarantees a shared baseline performance across all the plant sets is embedded in the learning process, so that the agent learns only the residual compensation for each uncertainty, thereby substantially enhancing sample efficiency. The proposed framework is applied to the design of an active vibration controller for automotive powertrains as a practical industrial application. Comparative validation demonstrates that the resulting controller remains robust against structural nonlinearities and dynamic variations over a wide range of plant conditions while improving the control performance.

**Keywords**: Active vibration control, Domain randomization, Continual learning, Model-based control, Robust control, Deep reinforcement learning

# 1. Introduction

## 1.1. Motivation

In modern industrial applications, including automotive powertrain systems [1], robotic platforms [2][3] and flexible structures [4], performance demands have become increasingly stringent, resulting in considerably higher system complexity. Such mechanical systems commonly exhibit nonlinear behaviors [5][6], communication delays, and uncertainties induced by variations in the system parameters [7]. Consequently, control strategies must be designed to simultaneously address these multiple sources of uncertainty in an integrated manner in order to achieve reliable performance.

Model-based control has achieved tremendous success across a wide range of mechanical systems; however, it fundamentally relies on the assumption that accurate and complete models of real-world systems are available. In practice, this assumption is rarely satisfied, and performance degradation caused by discrepancies between plant models and real systems is widely recognized as the *robust control problem* in control theory and as the *sim-to-real gap* in the machine learning community. Conventional robust control methods, such as $H_\infty$ control [8], guarantee robustness through a single worst-case design over a bounded uncertainty set. However, for robotic and automotive systems in which parameter variations and strong nonlinearities are intricately intertwined, covering all the heterogeneous uncertainties with a single linear controller inevitably yields overly conservative performance, and structural nonlinearities such as backlash violate the underlying design assumptions themselves. Meanwhile, rapid advances in computational resources have driven remarkable progress in artificial intelligence, which has recently demonstrated significant potential as an alternative control paradigm through numerous industrial applications. In particular, deep reinforcement learning (DRL) [9][10], emerging from the integration of deep neural networks (DNNs) and reinforcement learning (RL) [11], has attracted considerable attention [12][13][14]. A growing body of related work has shown that DRL can learn practically effective control policies for nonlinear, complex, large-scale, and high-dimensional plants—such as robotic systems [15], unmanned aerial vehicles [16], and powertrain control problems [17]—without relying on explicit system models. However, as DRL is based on trial-and-error interactions with the environment, learning directly in real-world systems is inherently dangerous [18]. Moreover, collecting the massive amounts of training data required through repeated real-world experiments is often impractical [19].

## 1.2. Related work

The success of DRL is largely attributed to its model-free nature and the generalization capability of DNNs. In recent years, increasing attention has been paid to training in simulation environments using domain randomization (DR) [18][20][21]. In simulation, where safety concerns are eliminated and virtually unlimited training data can be generated, DR intentionally injects random variations into the parameters of the simulation dynamics during training. Intuitively, by exposing an agent to a wide range of plant dynamics and encouraging it to learn policies that perform well across such variations, robustness against real-world systems can be enhanced. Consequently, DR has been widely adopted as an effective sim-to-real transfer for complex systems in which modeling the state transition dynamics is challenging, such as robotics control [22], locomotion tasks [23], and humanoid robots [24]. Nevertheless, when the training environment simultaneously involves multiple nonlinear characteristics and parameter variations, DR is known to produce excessively conservative and sub-optimal policies [21]. This is because excessive randomization across many dynamic factors increases the uncertainty perceived by the agent and exacerbates the task complexity, thereby making learning more difficult and time-consuming.

Several approaches have been proposed to address this challenge. Active domain randomization [25] identifies the most informative regions of the parameter space instead of sampling it uniformly, whereas automatic domain randomization [26] gradually widens the randomization strength as the policy improves. However, when a plant simultaneously exhibits multiple and diverse sources of uncertainty, these approaches alone may not be sufficient.

A natural remedy for this difficulty is to introduce the sources of uncertainty sequentially, rather than randomizing all of them simultaneously, so that the policy can acquire the strategy for each uncertainty step by step. However, such sequential training exposes neural networks to *catastrophic forgetting*, in which the knowledge acquired for earlier uncertainties is overwritten during subsequent training. Continual learning (CL) directly addresses this difficulty by enabling the accumulation of knowledge across a sequence of distinct tasks [27][28][29]. Existing CL approaches can be broadly categorized into several classes. Regularization-based methods [27] introduce additional constraints into the objective function to prevent parameters that are important for previously learned tasks from being excessively updated during the learning of new tasks. Expansion-based methods [30] dynamically extend the network architecture as new tasks are introduced, whereas replay-based methods [31] retain a small subset of data from past tasks in a replay buffer and repeatedly reuse it during subsequent training phases. More recently, these ideas have been further extended to the more challenging setting of continual RL [32][33]. Despite these advances, CL remains vulnerable to catastrophic forgetting when the number of tasks increases or when the discrepancy between tasks becomes pronounced [34][35]. Increasing the model capacity can mitigate forgetting to some extent [36]; however, this strategy is often impractical owing to the associated computational and training costs [34][37]. Among empirical studies on robustness against forgetting, a notable insight is the effectiveness of the *pretrain-then-finetune* paradigm in large-scale language models and image classification networks [38]. This observation implies the importance of first establishing a shared baseline level of performance that is transferable across tasks, prior to task-specific adaptation. A natural question that arises here is how this baseline performance (i.e., nominal performance) is defined within the framework of robust control theory.

### 1.3. Contribution and novelty

This study proposes a novel learning framework for acquiring robust control policies for complex controlled plants in which nonlinear characteristics and parameter variations are intricately intertwined. Attempting to address all the sources of uncertainty simultaneously within a single training process, as in previous studies on conventional DR-based training [17][18], often constitutes a fundamental cause of failure. A natural alternative is to progressively accumulate learning in a sequential manner, while retaining knowledge from previously learned policies. The proposed learning approach is outlined in Fig. 1. It is based on two key ideas as follows.

The first key idea is to decompose the original control problem according to individual sources of uncertainty and to view the learning process as continual learning, in which the strategies for handling each uncertainty are acquired sequentially. In contrast to a related work [21], which sequentially randomizes simulator parameters, we newly formulate the task sequence from the robust control perspective—as a plant set gradually encompassing structural nonlinearities as well as parameter variations. Rather than addressing all the nonlinearities and parameter variations simultaneously, a robust policy can be obtained more effectively through learning across a sequence of tasks. By incrementally enlarging the plant set, i.e., by gradually increasing the diversity and complexity of the learning tasks, policies capable of compensating for each source of uncertainty are acquired in a progressive manner. The knowledge obtained in each task is stably accumulated through CL without inducing abrupt changes in the policy, i.e., catastrophic forgetting.

The second idea is to incorporate a physical model-based controller (MBC), which guarantees a shared baseline performance throughout the entire task sequence (i.e., all the plant sets), into the learning process in order to accelerate the convergence of the residual policy. Unlike existing residual RL [39][40][41] designed for a single fixed task, the residual policy itself is continually updated across the whole task sequence without forgetting. This strategy contributes toward improved sample efficiency and accelerated learning.

In the resulting architecture, each constituent plays a complementary and indispensable role: the curriculum decomposition renders the multi-uncertainty problem tractable, CL preserves the strategies acquired for previous uncertainties, and the MBC-based residual structure supplies the shared baseline that keeps the sequential learning efficient.

In summary, the contributions and technical novelties of this study are as follows:

1. A new curriculum-based continual learning framework, referred to as *continual uncertainty learning* (CUL), is formulated for nonlinear systems on which multiple heterogeneous uncertainties are simultaneously superimposed. From the robust control perspective, the original system is extended into a finite set of plants whose uncertainties, including structural nonlinearities, are progressively expanded, and the acquisition of a robust policy is defined as an optimization problem with respect to the average performance over this plant set. The policy adaptation across the entire plant set is thereby formulated as a continual learning process, in which the policy is updated sequentially for each task (i.e., each uncertainty) with a memory-efficient anti-forgetting mechanism.

2. As the task sequence lengthens and the discrepancy between consecutive tasks grows, sequential learning alone becomes sample-inefficient. This difficulty is removed by embedding an MBC that provides a shared baseline performance across all the plant sets, allowing the DRL agent to focus on compensating for the residual gap on top of it.
3. The framework is applied to the design of an active vibration control system for automotive powertrains as a practical industrial application. Comparative validation against ablated variants as well as state-of-the-art learning-based approaches demonstrates that the resulting policy is more robust against structural nonlinearities and dynamic variations over a wide range of plant conditions.

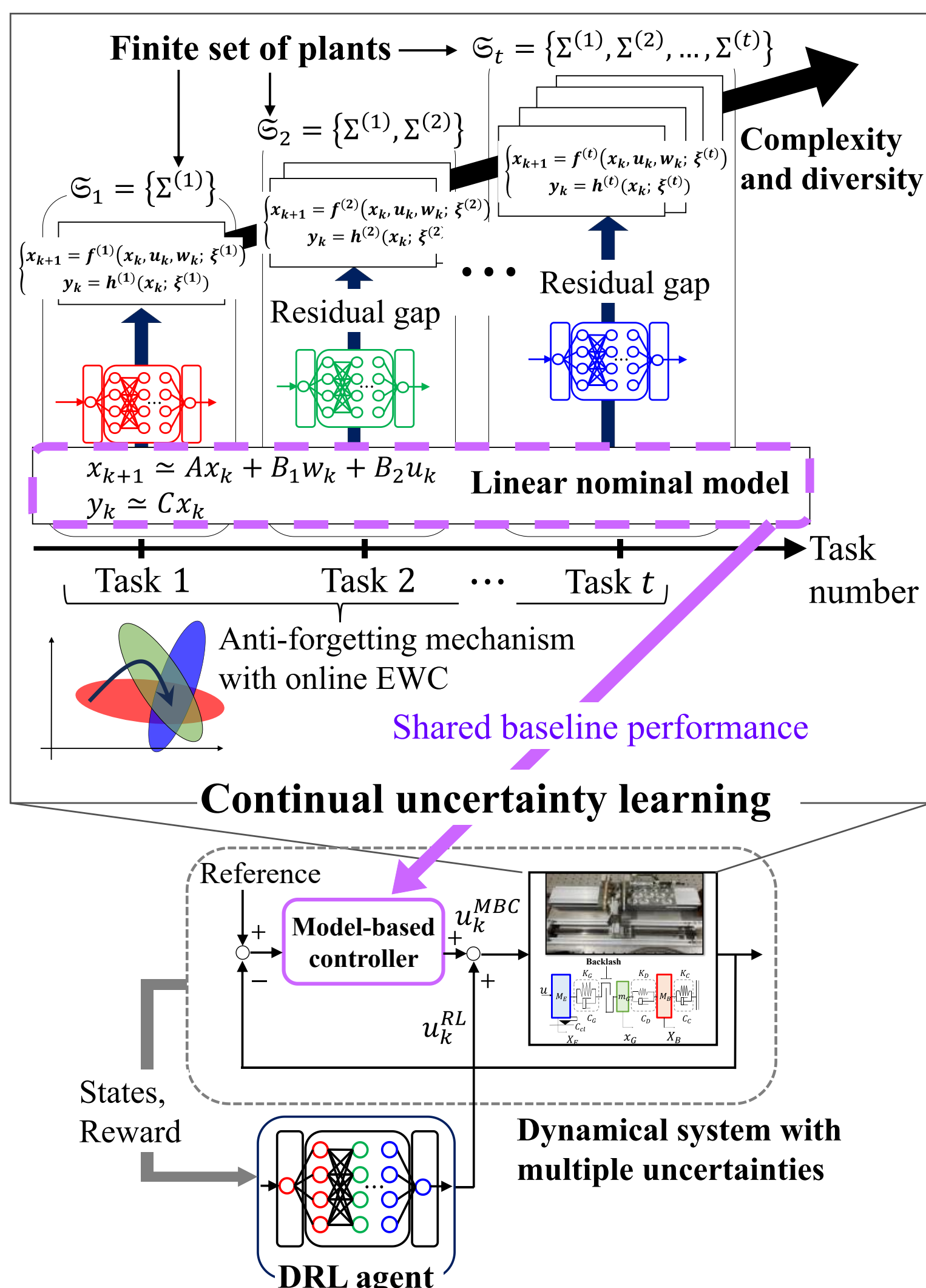

Fig. 1 Outline of the proposed approach.

The remainder of this paper is organized as follows. First, Section 2 defines the problem. Then, Section 3 reviews standard DRL setups. Section 4 describes the proposed method in detail. Section 5 investigates the performance of the proposed method through numerical experiments involving an uncertain powertrain model, with a focus on active vibration control and robustness verification. Finally, Section 6 summarizes the findings.

## 2. Problem formulation

We consider a discrete-time nonlinear system described by

$$x_{k+1} = f(x_k, u_k, w_k;\ \xi) \tag{1}$$

$$y_k = h(x_k;\ \xi) \tag{2}$$

where $x_k \in \mathbb{R}^n$, $u_k \in \mathbb{R}^m$, and $w_k \in \mathbb{R}^l$ denote the system state, control input, and external disturbance at time step $k$, respectively, and $y_k \in \mathbb{R}^o$ represents the controlled output. The nonlinear mapping $f(\cdot): \mathbb{R}^n \times \mathbb{R}^m \times \mathbb{R}^l \to \mathbb{R}^n$ characterizes the state transition dynamics, while $h(\cdot): \mathbb{R}^n \to \mathbb{R}^o$ defines the system output. Both functions are subject to uncertainties. The symbol $\xi = \{\xi_1, \xi_2, \cdots, \xi_N\}$ denotes a set of all the system uncertainties arising from parameter variations and nonlinearities, where each uncertainty has an internal parameter space $\xi_i \in \Xi_i$. The control objective is to determine the control input sequence $u_k$ such that the tracking error $e_k := y_k^r - y_k$ asymptotically converges to zero, i.e., $\lim_{k \to \infty} \|e_k\| = 0$, where $y_k^r \in \mathbb{R}^o$ denotes a given reference signal.

Under a control policy $u_k = \pi(x_k)$, the associated finite-horizon optimal control problem over a time horizon $T$ is formulated using an immediate cost function $c: \mathbb{R}^n \times \mathbb{R}^m \to \mathbb{R}$:

$$\min_{\{u_0, u_1, \cdots u_T\} \in U} \sum_{k=0}^{T} c(x_k, u_k) \tag{3}$$

subject to $x_{k+1} = f(x_k, u_k, w_k;\ \xi)\ \forall k \in [0, T]$

where $c(x_k, u_k)$ is a positive-definite stage cost. In this paper, we employ the quadratic stage cost:

$$c(x_k, u_k) = {e_k}^T Q e_k + {u_k}^T R u_k \tag{4}$$

where $Q \in \mathbb{R}^{o \times o}$ and $R \in \mathbb{R}^{m \times m}$ are positive semi-definite and positive definite symmetric matrices, respectively.

From the perspective of robust control, the original system can be viewed as an uncertain system in which the linear nominal model is subject to multiple additive uncertainties. For linear models, optimal model-based controllers can be easily designed, providing the baseline performance common to each task in CL. For most industrial systems, partially known system dynamics are available to derive the linearized nominal model. The following assumptions are introduced in this study.

***Assumption***: We have the linearized nominal model for the nonlinear system in Eqs. (1) and (2):

$$x_{k+1} = A x_k + B_1 w_k + B_2 u_k \tag{5}$$

$$y_k = C x_k \tag{6}$$

where $A \in \mathbb{R}^{n \times n}$, $B_1 \in \mathbb{R}^{n \times l}$, $B_2 \in \mathbb{R}^{n \times m}$, and $C \in \mathbb{R}^{o \times n}$. Several methods, including data-driven methods, system identification, and first principle-based modeling, can provide these system matrices. The model-based control input $u_k^{MBC} \in \mathbb{R}^m$ can be generated by designing a model-based linear controller for the above linear approximation model in the form of the following discrete state-space representation.

$$x_{k+1}^{c} = A_c x_k^c + B_c e_k \quad (7)$$

$$u_k^{MBC} = C_c x_k^c + D_c e_k \quad (8)$$

where $A_c \in \mathbb{R}^{n \times n}$, $B_c \in \mathbb{R}^{n \times o}$, $C_c \in \mathbb{R}^{m \times n}$ and $D_c \in \mathbb{R}^{m \times o}$. Here, $x_k^c \in \mathbb{R}^n$ is an internal state vector of MBC.

## 3. Deep reinforcement learning

### 3.1. Markov decision process

This section follows the RL framework based on descriptions in the literature [21][42][43]. We consider a RL formulation based on a Markov decision process (MDP), defined as $\mathcal{M} = (\mathcal{S}, \mathcal{A}, \mathcal{P}, \mathcal{R}, \rho_0, \gamma, T)$. Here, $\mathcal{S} = \mathbb{R}^n$ denotes the state space and $\mathcal{A} = \mathbb{R}^m$ denotes the action space, while the stationary transition dynamics distribution $\mathcal{P}(s_{k+1} \in \mathcal{S} \mid s_k \in \mathcal{S}, a_k \in \mathcal{A})$ governs the probabilistic evolution of the system. The real-valued scalar function $r_k = r(s_k, a_k) \in \mathcal{R}: \mathcal{S} \times \mathcal{A} \to \mathbb{R}$ specifies the immediate scalar-valued reward, $\rho_0$ characterizes the distribution of initial states, $\gamma \in (0,1]$ is the discount factor, and $T$ is the finite horizon for each episode. At each time step $k$, an agent receives a state $s_k$, takes an action $a_k$ and receives a scalar-valued reward $r_k$ by interacting with a stochastic environment $E$ in order to maximize a cumulative long-term reward. A parametric stochastic policy $\pi_\theta(a_k \mid s_k): \mathcal{S} \to \mathcal{P}(\mathcal{A})$, where $\mathcal{P}(\mathcal{A})$ is a set of probability measures on $\mathcal{A}$ and $\theta \in \mathbb{R}^{d1}$ is a parameter vector of $d1$ dimensions, is used to produce control actions. The RL objective is to identify the parameters $\theta$ that maximize the expected cumulative discounted reward, i.e., action-value function:

$$Q^{\pi_\theta}(s, a) = \mathbb{E}_{\pi_\theta}\left[\sum_{k=0}^{T} \gamma^k r(s_k, a_k) \mid S_0 = s, A_0 = a; \pi_\theta\right] \quad (9)$$

where the states and actions evolve according to $s_0 \sim \rho_0$, $a_k \sim \pi_\theta(\cdot \mid s_k)$, and $s_{k+1} \sim \mathcal{P}(\cdot \mid s_k, a_k)$. The discounted state visitation distribution for a policy $\pi$ is denoted as $\rho^\pi$.

### 3.2. Deep deterministic policy gradient

To solve the control problem under continuous state–action spaces, we employ the DDPG algorithm [42][43]. DDPG is an actor–critic method that learns a deterministic policy and uses an auxiliary critic network to approximate the action-value function.

Let $\mu_\theta(s_k)$ denote a deterministic policy parameterized by $\theta \in \mathbb{R}^{d1}$, i.e.,

$$a_k = \mu_\theta(s_k). \quad (10)$$

During training, exploration is encouraged by injecting noise into the deterministic policy:

$$\beta(s_k) = \mu_\theta(s_k) + \mathcal{N}_k, \quad (11)$$

$$a_k \sim \beta(s_k) \quad (12)$$

where $\mathcal{N}_k$ is sampled from a suitable noise process and $\beta(\cdot)$ denotes a stochastic exploration policy.

In place of the true action-value function $Q^{\mu_\theta}(s,a)$, we consider a DNN function approximator $Q^{\phi}(s,a)$, which is the critic network parameterized by $\phi \in \mathbb{R}^{d2}$. The critic is updated by minimizing the temporal-difference loss:

$$L(\phi) = \mathbb{E}_{s_k \sim \rho^\beta, a_k \sim \beta, r_k \sim E}\left[\left(Q^{\phi}(s_k, a_k) - Y_k\right)^2\right] \tag{13}$$

where the target value is given by

$$Y_k = r(s_k, a_k) + \gamma Q^{\phi_{tn}}\left(s_{k+1}, \mu_{\theta_{tn}}(s_{k+1})\right). \tag{14}$$

Here, $\phi_{tn}$ and $\theta_{tn}$ denote slowly updated target-network parameters as $\phi_{tn} \leftarrow \eta\phi + (1-\eta)\phi_{tn}$ and $\theta_{tn} \leftarrow \eta\theta + (1-\eta)\theta_{tn}$ with $\eta \ll 1$, respectively. The slow updating strategy improves the stability of the training process [43].

The actor parameters $\theta$ are updated in the direction of the gradient of the critic's estimate with respect to the actor parameters:

$$\begin{aligned} \nabla_\theta J(\mu_\theta) &\approx \mathbb{E}_{s_k \sim \rho^\beta}\left[\nabla_\theta Q^{\phi}(s,a)\big|_{s=s_k, a=\mu_\theta(s_k)}\right] \\ &= \mathbb{E}_{s_k \sim \rho^\beta}\left[\nabla_a Q^{\phi}(s,a)\big|_{s=s_k, a=\mu_\theta(s_k)} \nabla_\theta \mu_\theta(s)\big|_{s=s_k}\right]. \end{aligned} \tag{15}$$

where $\mathbb{E}_{s_k \sim \rho^\beta}[\cdot]$ is taken over the off-policy state distribution $\rho^\beta$ as the update direction varies for each visited state.

## 4. Proposed approach

### 4.1. Continual uncertainty learning

For the system with multiple uncertainties $\xi$ in Eqs. (1) and (2), we newly consider a set of nonlinear dynamical systems, each of which has its own structural uncertainties such as parameter variations and nonlinear characteristics. We then define progressive expansion of the plant set for continual learning. The original plant dynamics in Eqs. (1) and (2) depend on a set of $N$ uncertainty components,

$$\xi = \{\xi_1, \xi_2, \dots, \xi_N\}, \tag{16}$$

where each $\xi_i \in \Xi_i$ represents a distinct source of model uncertainty such as parameter variations or nonlinear effects.

Training a controller directly on the plant containing all the components of $\xi$ simultaneously can be difficult owing to the large variability induced by the combined uncertainties. To mitigate this difficulty, we propose a stage-wise *continual uncertainty learning* procedure in which the number of active uncertainty components is gradually increased. At training stage $t$, we define the active uncertainty set as

$$\xi^{(t)} := \{\xi_1, \xi_2, \dots, \xi_t\}, \qquad t = 1,2,\dots,N. \tag{17}$$

As $t$ increases, the set of uncertainties $\xi^{(t)}$ is expanded in the sense that

$$\xi^{(1)} \subset \xi^{(2)} \subset \cdots \subset \xi^{(N)} = \xi, \tag{18}$$

where each $\xi^{(t)}$ contains all the uncertainty components introduced in the previous training stages and one additional component $\xi_t$. The final stage recovers the original plant with the full uncertainty set $\xi$. This construction ensures that the complexity of the plant set used for CL increases gradually as learning progresses.

Correspondingly, the plant dynamics added at training stage $t$ are denoted as

$$\Sigma^{(t)}: \begin{cases} x_{k+1} = f^{(t)}\left(x_k, u_k, w_k;\ \xi^{(t)}\right) \\ y_k = h^{(t)}(x_k;\ \xi^{(t)}) \end{cases}, \tag{19}$$

where $f^{(t)}$ and $h^{(t)}$ denote the system mappings in which only the uncertainty components $\xi_1, \dots, \xi_t$ are treated as variable, while the remaining components are fixed to nominal values.

The corresponding set of plants available for training at stage $t$ is defined as

$$\mathfrak{S}_t := \left\{\Sigma^{(1)}, \Sigma^{(2)}, \dots, \Sigma^{(t)}\right\}. \tag{20}$$

To formalize the idea that the control policy is trained on gradually more diverse and complex plants, we introduce an increasing sequence of plant sets $\mathfrak{S}_t$, indexed by the task training stage $t$:

$$\mathfrak{S}_0 \subset \mathfrak{S}_1 \subset \mathfrak{S}_2 \subset \mathfrak{S}_3 \subset \cdots \subset \mathfrak{S}_N. \tag{21}$$

This sequence naturally represents a curriculum in which the learning algorithm encounters increasingly complex plants as the training advances. The goal of this study is to design a learning-based controller that realizes desirable control performance over the entire set $\mathfrak{S}_N$. During the training process, a single new plant is added one at a time at each stage according to

$$\mathfrak{S}_{t+1} = \mathfrak{S}_t \cup \left\{\Sigma^{(t+1)}\right\}. \tag{22}$$

This formulation ensures that the training environment expands step-by-step as the learning progresses, rather than incorporating a batch of new plants simultaneously. In the final stage, the set of uncertainties becomes $\xi^{(N)} = \xi$, and the plant $\Sigma^{(N)}$ represents the original system with all the uncertainty components active. Exceptionally, $\mathfrak{S}_0 = \{\Sigma^{(0)}\}$ denotes an idealized plant with all the uncertainties disabled, i.e., the linearized nominal system in Eqs. (5) and (6).

At training stage $t$, the control policy $\pi_\theta^{(t)}$ is optimized over the plant set $\mathfrak{S}_t = \{\Sigma^{(i)} \mid i \in \{1,2, \dots, t\}\}$. The corresponding learning problem is given as

$$\begin{aligned} &\min_{\pi_\theta^{(t)}} \mathbb{E}_{i\sim p_t(i)}\left[J_i\left(\pi_\theta^{(t)}\right)\right] \\ &\text{subject to } \Sigma^{(i)} \in \mathfrak{S}_t, \end{aligned} \tag{23}$$

where $p_t(i)$ denotes a task probability distribution defined over $\mathfrak{S}_t$ (e.g., uniform distribution), and $J_i(\pi_\theta^{(t)})$ is the cumulative cost of policy $\pi_\theta^{(t)}$ when applied to plant $\Sigma^{(i)}$. This formulation allows the controller to be trained on the subset of plants available at stage $t$, while progressively adapting to increasing variability in the subsequent stages. We aim to find the optimal global policy that minimizes the expected performance with respect to probability distribution $p_N(i)$ across all the control tasks.

Solving the optimal control problem in Eq. (23) over $\mathfrak{S}_N$ can be interpreted, from the viewpoint of machine learning, as a CL problem involving multiple tasks. Here, the control of each plant— corresponding to a specific type of uncertainty—is regarded as an individual task. More specifically, the sequential policy update should be formulated as the solution to a regularized optimization problem:

$$\pi_\theta^{(t+1)} = \arg\min_\pi [\mathbb{E}_{i\sim p_{t+1}(i)}[J_i(\pi)] + \lambda\,\Omega(\pi, \pi_\theta^{(t)})] \tag{24}$$

where $\Omega(\pi, \pi_\theta^{(t)})$ is a regularization term that penalizes large abrupt deviations from the previous policy (e.g., weight-distance or Kullback–Leibler divergence), and $\lambda > 0$ is a coefficient that denotes a trade-off between learning a new task and not forgetting already-experienced old tasks. In a sequence of consecutive tasks, a newly introduced plant $\Sigma^{(t+1)}$ at a later stage of Eq. (22) may differ significantly from those previously used for training, resulting in abrupt discrepancies between tasks. The regularization term $\Omega(\pi, \pi_\theta^{(t)})$ in Eq. (24) suppresses excessive policy changes induced by such discrepancies, thereby preventing overfitting to the newly introduced plant at the expense of forgetting the previously learned ones.

### 4.2. Memory-efficient anti-forgetting mechanism

This study employs EWC, a regularization-based CL algorithm inspired by task-specific synaptic consolidation in the human brain [27]. EWC alleviates catastrophic forgetting by selectively slowing down updates of parameters that are deemed important for past tasks, thereby preserving the performance achieved on those tasks without inducing forgetting. The importance of each parameter is quantified using the Fisher information matrix (FIM) [27]. Accordingly, the loss function to be minimized during the training of the actor network is augmented with a weighted quadratic penalty based on the parameter importance for past tasks:

$$\mathcal{L}(\theta_t) = \mathcal{L}_t(\theta_t) + \sum_{m=1}^{t-1} \frac{\lambda}{2} \|\theta_t - \theta_m^*\|_{F_m}^2 = \mathcal{L}_t(\theta_t) + \sum_{m=1}^{t-1} \sum_j \frac{\lambda}{2} F_{m,j} \left(\theta_{t,j} - \theta_{m,j}^*\right)^2 \tag{25}$$

where $j$ denotes the index of each parameter. Here, $\theta_t \in \mathbb{R}^{d1}$ represents the parameter vector to be optimized for the current $t$-th task, and $\theta_{t,j}$ is its $j$-th component. The overall loss function $\mathcal{L}(\theta_t)$ consists of the task-specific loss $\mathcal{L}_t(\theta_t)$ for the current task and a regularization term that penalizes excessive deviations from parameters $\theta_m^* \in \mathbb{R}^{d1}$ optimized for previous tasks. The importance of each parameter $\theta_{m,j}^*$ for the $m$-th task is quantified by $F_{m,j}$, which corresponds to the diagonal elements of FIM $F_m \in \mathbb{R}^{d1\times d1}$ computed for each task.

The diagonal empirical FIM $F(\theta) \in \mathbb{R}^{d1\times d1}$ for the latest task is computed using the squared gradients of the log-likelihoods of the action predicted by the actor policy $\pi_\theta$ [21]:

$$\begin{aligned} F(\theta) &= \mathbb{E}_{x,y}[\nabla_\theta \log \pi_\theta(y \mid x)\, \nabla_\theta \log \pi_\theta(y \mid x)^T] \\ &\approx \frac{1}{n} \sum_{l=1}^{n} [\nabla_\theta \log \pi_\theta(y_l \mid x_l)\, \nabla_\theta \log \pi_\theta(y_l \mid x_l)^T] \end{aligned} \tag{26}$$

where $x_l$ and $y_l$ denote the input and output of the policy network, and the number of samples in the replay buffer is represented as $n$.

However, standard EWC requires storing the FIMs and the optimal parameters for all the previously learned tasks (i.e., plant uncertainties), which leads to substantial storage requirements. To alleviate this issue, this study proposes the integration of online-EWC [44] with DDPG. Specifically, only the optimal parameters $\theta_{t-1}^*$ and the corresponding online FIM $F_{t-1}^*$ from the most recent task are retained, and the following loss function is minimized:

$$\mathcal{L}(\theta_t) = \mathcal{L}_t(\theta_t) + \frac{\lambda}{2} \| \theta_t - \theta_{t-1}^* \|_{\gamma F_{t-1}^*}^2 . \tag{27}$$

The online FIM $F_t^*$ is updated using the Fisher information of the current task $F_t$ and a hyperparameter $\gamma$:

$$F_t^* = \gamma F_{t-1}^* + F_t . \tag{28}$$

Here, the elements of the online FIM are normalized by their maximum value after each task, which decouples the fixed coefficient $\lambda$ from the task-dependent scale of the FIM in a similar spirit to [21]. As DDPG adopts a deterministic policy, it is common practice to define the FIM through the policy Jacobian [45][46], which corresponds to assuming a Gaussian policy with a fixed variance. Accordingly, the parameter importance is computed as:

$$F(\theta) \approx \frac{1}{n} \sum_{l=1}^{n} [\nabla_\theta \mu_\theta(s_l) \nabla_\theta \mu_\theta(s_l)^T], \tag{29}$$

$$F_j = \frac{1}{n} \sum_{s_l \in \text{buffer}} \left( \frac{\partial \mu_\theta(s_l)}{\partial \theta_j} \right)^2 , \tag{30}$$

where the importance of each parameter $\theta_j$ for the previous task is measured by $F_j$, which corresponds to the diagonal elements of $F(\theta)$. In the actor–critic architecture, the anti-forgetting regularization is applied only to the actor network, which directly determines the control input, whereas the critic serves as an auxiliary estimator refreshed for each task [21][46]; this also preserves adaptability to newly introduced uncertainties.

### 3.3. Domain randomization over the plant set

To enhance the generalization capability of the actor policy, this study consistently applies DR throughout the CL process. Specifically, for the environment model $\Sigma^{(i)}$ selected at each episode, the uncertain dynamics parameters $\xi^{(i)}$ are randomly sampled according to a probability distribution $\rho_\xi^{(i)}$. Accordingly, for each episode, the environment dynamics are determined by a realization of the uncertainty parameters $\xi^{(i)}$ sampled from $\rho_\xi^{(i)}$, so that training is performed over a family of dynamics induced by domain randomization [18][20][47]. Under this formulation, the objective function can be defined on the basis of the expected return over the distribution of the dynamics model:

$$J_i(\pi_\theta) = -\mathbb{E}_{\xi^{(i)}\sim\rho_\xi^{(i)}}\left[\mathbb{E}_{\tau^{(i)}\sim p(\tau^{(i)}|\pi_\theta,\xi^{(i)})}\left[\sum_{k=1}^{T}\gamma^{k-1}r(s_k,a_k)\right] \mid s_1,a_1;\pi_\theta\right] \tag{31}$$

where the likelihood of a trajectory $\tau^{(i)} = (s_1, a_1, s_2, \dots, a_{T-1}, s_T)^{(i)}$ is denoted by $p(\tau^{(i)} \mid \pi_\theta, \xi^{(i)})$ under the policy $\pi_\theta$. Therefore, the learning problem in Eq. (24) is more specifically formulated as follows:

$$\begin{aligned}
\pi_\theta^{(t+1)} &= \arg\min_{\pi}[\mathbb{E}_{i\sim p_{t+1}(i)}[J_i(\pi)] + \lambda\,\Omega(\pi,\pi_\theta^{(t)})] \\
&= \arg\min_{\pi}\Bigg[-\mathbb{E}_{i\sim p_{t+1}(i)}\left[\mathbb{E}_{\xi^{(i)}\sim\rho_\xi^{(i)}}\left[\mathbb{E}_{\tau^{(i)}\sim p(\tau^{(i)}|\pi,\xi^{(i)})}\left[\sum_{k=1}^{T}\gamma^{k-1}r(s_k,a_k)\right] \mid s_1,a_1;\pi\right]\right] \\
&\quad + \lambda\,\Omega(\pi,\pi_\theta^{(t)})\Bigg]
\end{aligned} \tag{32}$$

### 4.4. Learning support by MBC

As the number of tasks increases or the discrepancy between consecutive tasks becomes pronounced, the learning efficiency of CL may deteriorate. In addition, solving the optimization problem in Eq. (32), which involves three nested expectation operators, requires a substantial number of samples. To mitigate this issue, this study integrates MBC into the training process, thereby accelerating learning progress and improving adaptation across tasks. This improvement arises from the fact that MBC in Eqs. (8) and (9) provides a shared baseline performance that captures the fundamental control behaviors common across plant sets $\mathfrak{S}_t$ $(t = 1,2,\dots,N)$. Consequently, MBC relieves the DRL agent from the need to learn the base control structure from scratch, thereby allowing it to focus instead on compensating for each residual gap between a baseline level of performance and the desired optimal behavior.

Accordingly, we define the control input based on a linear combination of the MBC input $u_k^{MBC}$ and the DRL agent's policy $u_k^{\mathrm{RL}} \leftarrow \pi_\theta$ as follows:

$$u_k = u_k^{MBC} + u_k^{RL} \tag{33}$$

In the above framework, the state transition of the training environment $E\left(\Sigma^{(i)}, \xi^{(i)}\right)$, which is randomly sampled at the beginning of each episode, satisfies the Markov property [11][48] as follows:

$$
\begin{aligned}
\mathcal{X}_{k+1}^{env} = \begin{bmatrix} x_{k+1} \\ x_{k+1}^{c} \end{bmatrix} &= \begin{bmatrix} f^{(i)}\left(x_k, u_k, w_k;\ \xi^{(i)}\right) \\ A_c x_k^c + B_c e_k \end{bmatrix} = \begin{bmatrix} f^{(i)}\left(x_k, u_k^{RL} + u_k^{MBC}, w_k;\ \xi^{(i)}\right) \\ A_c x_k^c + B_c e_k \end{bmatrix} \\
&= \begin{bmatrix} f^{(i)}\left(x_k, u_k^{RL} + C_c x_k^c + D_c (y_k^r - y_k), w_k;\ \xi^{(i)}\right) \\ A_c x_k^c + B_c (y_k^r - y_k) \end{bmatrix} \\
&= \begin{bmatrix} f^{(i)}\left(x_k, u_k^{RL} + C_c x_k^c + D_c y_k^r - D_c h^{(i)}(x_k;\ \xi^{(i)}), w_k;\ \xi^{(i)}\right) \\ A_c x_k^c + B_c y_k^r - B_c h^{(i)}(x_k;\ \xi^{(i)}) \end{bmatrix} \\
&= \mathcal{F}^{(i)}\left(\mathcal{X}_k^{env}, u_k^{RL}, w_k, y_k^r;\ \xi^{(i)}\right), \\
&\qquad \mathcal{F}^{(i)}(\cdot) \triangleq \begin{bmatrix} f^{(i)}(*) \\ A_c x_k^c + B_c e_k \end{bmatrix}
\end{aligned} \tag{34}
$$

where the state of the closed loop system is denoted by $\mathcal{X}_k^{env} \triangleq \left[{x_k}^T \quad {x_k^c}^T\right]^T$. Accordingly, the training environment retains the Markov property even under the mixed control input in Eq. (33).

The proposed learning algorithm is outlined in **Algorithm 1**.

**Algorithm 1:** The CUL Algorithm.

| | |
|---|---|
| **1** | Input: Environment $E$ and trajectories pool $\mathcal{T}_{rand}$ |
| **2** | Output: Optimal policy |
| **3** | Randomly initialize the critic parameters and the actor's policy parameters. |
| **4** | Design an MBC with a linearized nominal model $\mathfrak{S}_0 = \{\Sigma^{(0)}\}$. |
| **5** | **for** each task $t$ **do** |
| **6** | $\xi^{(t)} = \{\xi_1, \xi_2, \dots, \xi_t\}$ increase the active uncertainty set. |
| **7** | $\mathfrak{S}_t = \mathfrak{S}_{t-1} \cup \{\Sigma^{(t)}\}$ expand the set of plant dynamics. |
| **8** | **for** each episode **do** |
| **9** | $\Sigma^{(i)} \sim p_t(i)$ randomly select a plant from $\mathfrak{S}_t = \{\Sigma^{(i)} \mid i \in \{1,2,\dots,t\}\}$. |
| **10** | $\xi^{(i)} \sim \rho_\xi^{(i)}$ randomly sample dynamics for $\Sigma^{(i)}$ in DR. |
| **11** | $E \leftarrow \Sigma^{(i)}$ randomly selected environment. |
| **12** | Generate rollout $\tau^{(i)} = (s_1, a_1, \dots, s_T)^{(i)} \sim \pi_\theta^{(t)} + u_k^{MBC}$ with dynamics $\xi^{(i)}$ and $E$. |
| **13** | $\mathcal{T}_{rand} \leftarrow \mathcal{T}_{rand} \cup \tau^{(i)}$ |
| **14** | $u_k^{RL} \leftarrow \pi_\theta^{(t)}$ DRL control input from the actor policy $\pi_\theta^{(t)}$. |
| **15** | $u = u_k^{RL} + u_k^{MBC}$ combination of the DRL policy $u_k^{RL}$ and the model-based control $u_k^{MBC}$. |
| **16** | **for** each gradient step **do** |
| **17** | Update the critic parameters by minimizing the mean squared error. |
| **18** | Compute the actor loss with a weighted quadratic penalty based on the parameter importance. |
| **19** | Update the actor parameters using the sampled policy gradient. |
| **20** | $\theta_t \leftarrow \theta_t + \alpha \nabla_\theta J_i\left(\pi_\theta^{(t)}\right)$ with $\mathcal{T}_{rand}$ update. |
| **21** | $u = u_k^{RL} + u_k^{MBC}$ update the hybrid control. |
| **22** | **end** |
| **23** | **end** |
| **24** | Compute the Fisher information of the current task. |
| **25** | Save the snapshot of the optimal actor parameters $\theta_t$ of the current task. |
| **26** | **end** |

***Remark 1 (Closed-loop stability):*** The MBC guarantees nominal closed-loop stability, thus providing a stabilizing baseline. The actor's output is bounded by a saturating output layer; hence the actor policy acts as a bounded compensation around the nominally stabilized loop [40]. The extensive verifications presented later, including the Monte Carlo simulations in which no divergent behavior was observed, empirically support the stability of the resulting closed-loop system, while a rigorous Lyapunov-based guarantee remains future work.

## 5. Numerical verification

### 5.1. Active vibration control of a powertrain system with multiple uncertainties

This section evaluates the proposed continual learning framework through numerical verifications in which an active vibration control problem of an automotive powertrain system is simulated. The powertrain dynamics involve heterogeneous uncertainties, including parametric variations, operating-condition changes, and nonlinear effects, all of which are known to complicate vibration suppression [17]. The system overview is shown in Fig. 2.

A nonlinear automotive powertrain model [49][50] is considered, where the controlled output is the vehicle body vibration response $y_k = X_B$. The control objective is to attenuate transient vibrations by applying a control input to the actuator such that the tracking error $e_k = y_k^r - y_k$ converges to zero, as stated in Section 2. In addition to the nominal dynamics shown in Table 1, the system is subject to the following sources of uncertainty:

1. Mass variations in the vehicle body $M_B \in [M_B^{min}, M_B^{max}]$ and the actuator $M_E \in [M_E^{min}, M_E^{max}]$;
2. Damping coefficient variations in the drivetrain $C_G \in [C_G^{min}, C_G^{max}]$, $C_D \in [C_D^{min}, C_D^{max}]$, and $C_C \in [C_C^{min}, C_C^{max}]$;
3. Operating-condition changes, represented by variations in the reference signal $y_k^r$;
4. Nonlinear dynamics arising from mechanical backlash, which has various widths $\delta \in [\delta_{min}, \delta_{max}]$, in the drivetrain.

In particular, the backlash introduces a dead-zone nonlinearity that causes discontinuous mode switching between contact and non-contact phases, thereby increasing control difficulty. The coexistence of these uncertainties is the motivation for the adoption of a learning-based control framework capable of systematic adaptation under increasingly challenging conditions.

Table 1 Nominal parameters of the powertrain system.

| Parameter | Description | Nominal Value | Unit |
|---|---|---|---|
| $K_C$ | Spring connected with $M_B$ | 660.0 | N/m |
| $K_G$ | Spring between $M_E$ and $m_G$ | $5.3 \times 10^4$ | N/m |
| $K_D$ | Spring between $M_B$ and $m_G$ | $2.2 \times 10^4$ | N/m |
| $M_E$ | Mass of the actuator | 1.04 | kg |
| $m_G$ | Mass of the intermediate part | 0.039 | kg |
| $M_B$ | Mass of the vehicle body | 0.232 | kg |
| $C_D$ | Damper between $M_B$ and $m_G$ | 12.5 | Ns/m |
| $C_C$ | Damper connected with $M_B$ | 0.1 | Ns/m |
| $C_{cl}$ | Damper of $M_E$ | 1.5 | Ns/m |
| $C_G$ | Damper between $M_E$ and $m_G$ | 36.0 | Ns/m |
| $\delta$ | Length of the backlash | 0.005 | m |
| $y^r_{0-2s}$ | Steady value of the reference signal from 0 to 2 seconds | $-0.006$ | m |
| $y^r_{2-4s}$ | Steady value of the reference signal from 2 to 4 seconds | 0.0227 | m |

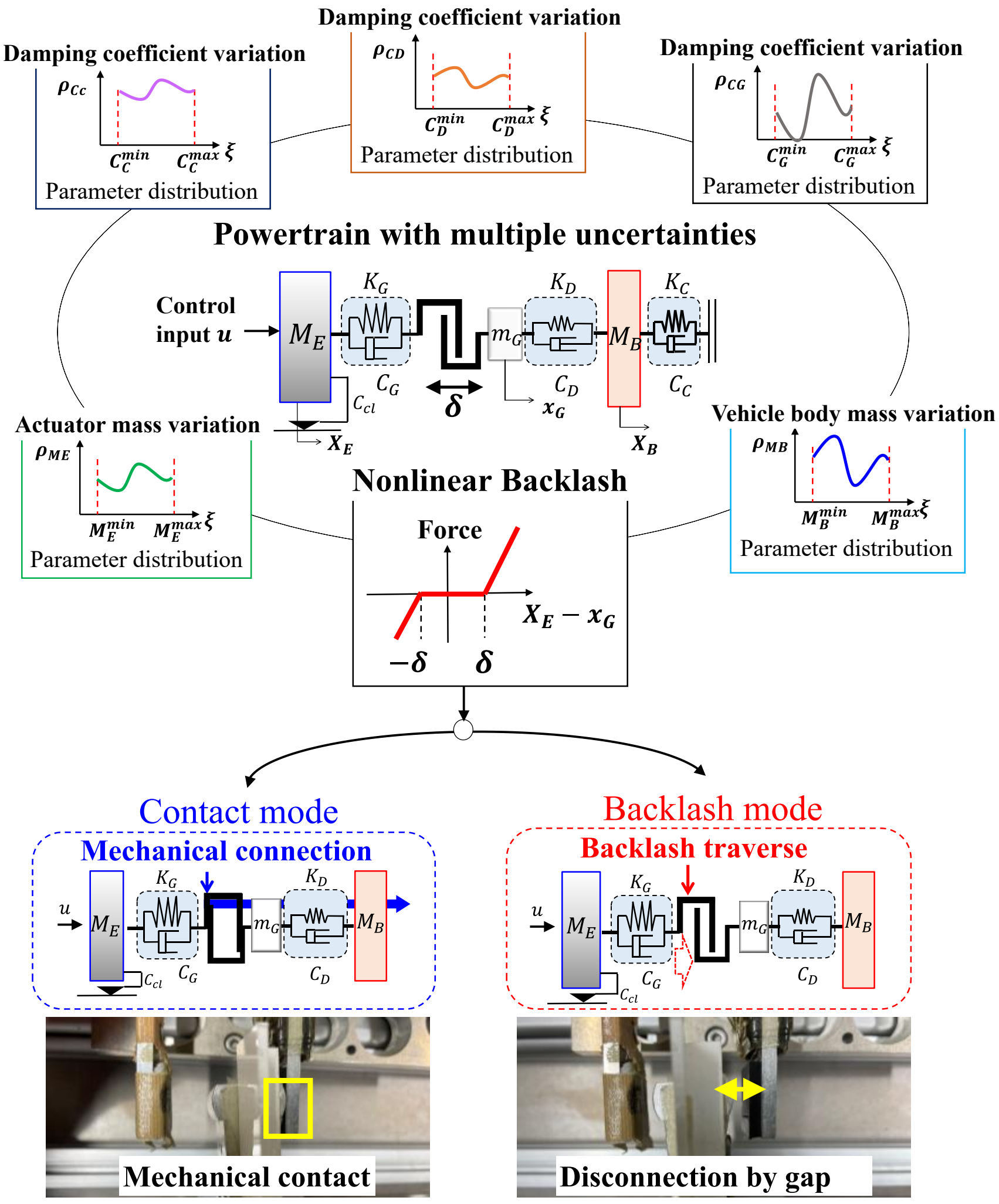

Fig. 2 Complex powertrain system with multiple uncertainties.

To design the MBC in Eqs. (7) and (8), a linearized nominal model of the powertrain system is employed, where backlash effects and parameter variations are neglected. Based on this nominal model, an output-feedback $H_2$ controller is designed using standard linear control theory [51][52]. For powertrain applications, a feedforward term is incorporated into the controller.

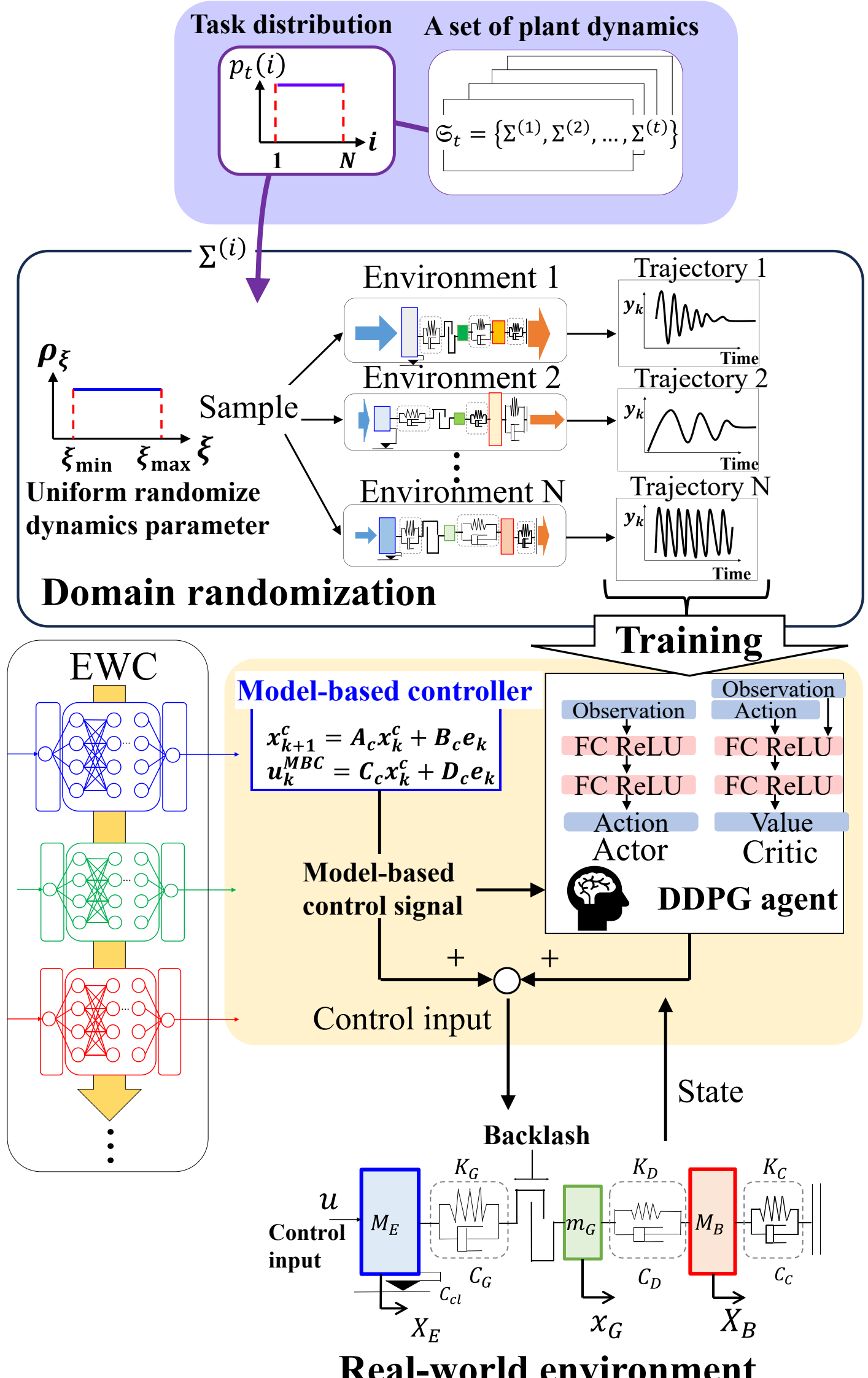

Fig. 3 Application of the proposed learning framework to active vibration control of the powertrain system.

## 5.2. CL tasks with progressive uncertainty expansion

To explicitly assess CL under progressively expanding uncertainties, a sequence of learning tasks is constructed by gradually expanding the set of plant uncertainties. Five tasks are defined as follows.

- **Task 0**:
  The linear nominal model is considered. All the parameters are fixed at their nominal values listed in Table 1. The reference signal is also fixed, and no backlash nonlinearity is present.
- **Task 1 ($t = 1$)**:
  In addition to Task 0, domain randomization is applied to the masses $M_B$ and $M_E$. The reference signal $y_k^r$ is randomly sampled at the beginning of each episode.
- **Task 2 ($t = 2$)**:
  Building upon Tasks 0 and 1, domain randomization is further applied to the damping coefficients $C_G$, $C_D$, and $C_C$, introducing additional parametric uncertainty in the drivetrain.
- **Task 3 ($t = 3$)**:
  A fixed-width backlash nonlinearity is introduced, structurally transforming the powertrain into a nonlinear system. The randomization range of the reference signal is also enlarged to increase operating-condition variability.
- **Task 4 ($t = 4$)**:
  In addition to all the uncertainties considered in Tasks 0–3, the backlash width itself is randomized, making the nonlinear characteristics subject to domain randomization.

Through Tasks 0–4, both the type and the extent of uncertainties are monotonically increased, resulting in a curriculum in which the training environment becomes progressively more diverse and complex. This task sequence allows systematic investigation of CL when robustness must be accumulated across heterogeneous uncertainties.

### 5.3. Verification settings

The RL agent observes the reference signal, the vehicle body vibration $X_B$, the tracking error, its integral and derivative, and the control input from MBC. All the observation components are normalized to facilitate stable learning. The reward is defined as the negative of a quadratic stage cost $r_k = -c(x_k, u_k)$ in Eq. (4) that penalizes tracking error and control effort.

For each task, the uncertain parameters associated with the corresponding uncertainty set are randomized at the beginning of every episode according to uniform distributions. The hyperparameters involved in the proposed algorithm are summarized in Table 2. The parameter variation ranges used for each uncertainty are listed in Table 3. All the training and evaluation procedures are conducted in MATLAB R2024b. The procedure is outlined in Fig. 3. The resulting control system consists of the linear MBC and a single forward pass through the actor network, and is therefore implementable within the sampling time listed in Table 2.

We implement multiple baselines for comparison, which include the counterparts of state-of-the-art learning-based approaches to the present control problem:

1. *No MBC*: CL (i.e., online-EWC-DDPG) is conducted without MBC. This configuration shares the underlying idea of a previous approach [21].

2. *Full randomization*: The policy is trained by activating all the uncertainties together at the same time from the beginning without any CL, while support by MBC is present, which corresponds to a previous approach in [17].
3. *Only MBC*: A model-based linear $H_2$ controller alone is directly applied to the powertrain system, which has no compensation for parametric variations and nonlinearity.

Table 2. Hyperparameters for the proposed algorithm.

| Hyperparameter | Value |
| --- | --- |
| Sampling time | $0.006$ s |
| Critic learning rate | $1.0 \times 10^{-4}$ |
| Actor learning rate | $1.0 \times 10^{-4}$ |
| Discount factor $\gamma$ | 0.99 |
| Exploration noise $\mathcal{N}$ | Ornstein-Uhlenbeck action noise |
| Number of neurons in the hidden layers | 128 |
| Size of the mini-batch $M$ | 128 |
| Size of the replay buffer | $1.0 \times 10^{5}$ |
| Target smooth factor $\eta$ | $1.0 \times 10^{-3}$ |
| Horizon $T$ | 667 |
| Network architecture | Fully connected + ReLU |
| Number of episodes | 500 |
| Number of episodes on each task | 100 |
| EWC reg. const. $\lambda$ | 1.0 |
| EWC replay batch $n$ | 128 |
| EWC replay samples | $1.0 \times 10^{5}$ |
| online-EWC $\gamma$ | 0.9 |

Table 3. Randomized parameters.

| Parameter | Range |
|---|---|
| Vehicle body mass: $M_B$ | Nominal value $\pm 50\%$: $\left[M_B^{min}, M_B^{max}\right] = [0.1160, 0.3480]$ |
| Actuator mass: $M_E$ | Nominal value $\pm 50\%$: $\left[M_E^{min}, M_E^{max}\right] = [0.5200, 1.5600]$ |
| Damping coefficient: $C_G$ | Nominal value $\pm 50\%$: $\left[C_G^{min},\ C_G^{max}\right] = [18, 54]$ |
| Damping coefficient: $C_D$ | Nominal value $\pm 50\%$: $\left[C_D^{min},\ C_D^{max}\right] = [6.25, 18.75]$ |
| Damping coefficient: $C_C$ | Nominal value $\pm 50\%$: $\left[C_C^{min},\ C_C^{max}\right] = [0.05, 0.15]$ |
| Length of backlash: $\delta$ (randomized only in *Scenario 2*) | Nominal value $\pm 50\%$: $[\delta_{min}, \delta_{max}] = [0.0025, 0.0075]$ |
| Steady value of the reference signal from 0 to 2 seconds: $y_{0-2s}^{r}$ | $\left[y_{0-2s}^{r\ min}, y_{0-2s}^{r\ max}\right] = [-0.01515, 0.030303\ ]$ m |
| Steady value of the reference signal from 2 to 4 seconds: $y_{2-4s}^{r}$ | $\left[y_{2-4s}^{r\ min}, y_{2-4s}^{r\ max}\right] = [-0.01515, 0.030303]$ m |

### 5.4. Results and discussion

Figure 4 shows the training curves in terms of episode rewards obtained by each DRL-based method. The CL task is switched every 100 episodes.

Compared with the proposed method (red line), *No MBC* (cyan line) requires a larger number of episodes to converge and exhibits an unstable learning process. In particular, a noticeable degradation in episode rewards is observed between 200 and 400 episodes, which suggests pronounced discrepancies between consecutive tasks. In the absence of MBC, which provides a baseline performance, the agent must learn control behaviors from scratch. Consequently, increased exploration is required, rendering the learning process more vulnerable to task switching. By contrast, the proposed method follows a stable learning trajectory without engaging in unnecessary exploration. The rapid convergence achieved with fewer episodes indicates that MBC effectively improves the learning efficiency in CL.

*Full randomization* shows immediate and significant reward fluctuations, while the proposed method's reward variance grows incrementally over episodes. This reflects the differing learning mechanisms, i.e., *full randomization* jointly incorporates all randomized uncertainties in all episodes.

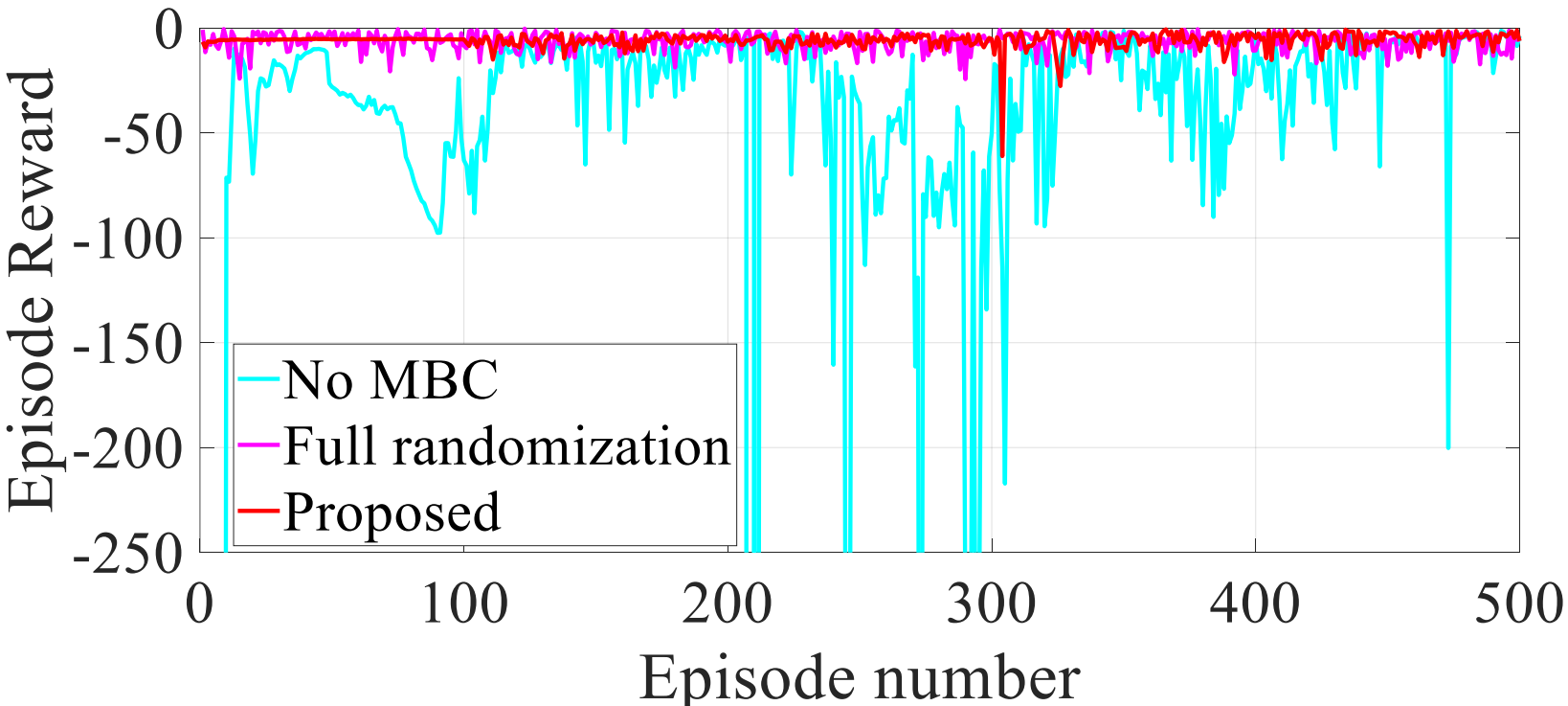

Fig. 4 Reward curve.

Figure 5 illustrates the control results for the linear nominal system, where the upper and lower plots represent the time histories of the vehicle body vibration and the control input, respectively. Table 4 provides a quantitative comparison of the considered methods by evaluating the 2-norm of the control error from the ideal reference response (green line in Fig. 5). All active control methods achieve sufficient vibration suppression compared with the open-loop response (blue line). The slight overshoot observed in the magenta line suggests a conservative policy induced by *full randomization*. According to Table 4, *only MBC* achieves the best performance. Since the plant considered in Fig. 5 contains no uncertainties and sim-to-real gap is absent, this result is theoretically reasonable.

The combinations of parameter variations are virtually infinite within the ranges listed in Table 3. Therefore, this study presents results for several representative cases, specifically those corresponding to verification conditions where each characteristic takes its maximum or minimum deviation. Figures 6–8 show the corresponding control results, and Tables 5–7 provide quantitative evaluations of Figs. 6–8 based on the 2-norm of the tracking error, respectively. In all verification cases, the proposed method achieves the smallest norm values, thereby demonstrating superior robustness.

In Fig. 6, in contrast to the proposed method (red line), which achieves satisfactory tracking performance, *No MBC* (cyan line) exhibits sustained vibrations. This result indicates that the presence or absence of MBC affects not only the efficiency of CL but also the robustness of the resulting policy. From Figs. 4 and 6, it can be concluded that the current number of episodes may be inadequate for solely leveraging DRL. In other words, without the baseline performance ensured by MBC, the sample efficiency of CL deteriorates. Although there exist conditions under which satisfactory control performance is achieved, as shown in Fig. 7 and Table 6, the effects of insufficient learning convergence become evident when the reference value—corresponding to driving conditions in real vehicles—changes drastically, as illustrated in Fig. 6. This can be attributed to the fact that learning with DRL alone does not provide extrapolation capability based on physical laws; consequently, when the discrepancies between tasks are pronounced, the agent is forced to reacquire baseline control action through trial-and-error. From this perspective, the comparison between the red and cyan lines demonstrates that MBC plays a crucial role in improving the learning efficiency.

Although *full randomization* maintains robustness under many conditions, residual vibrations and overshoot are observed in Figs. 7 and 8. This is due to the learning process in which all uncertainties are handled simultaneously within a single training stage. When numerous parameters are randomized from the outset in addition to nonlinear characteristics, the training environment becomes excessively uncertain, making it difficult for the DRL agent to acquire appropriate control behaviors. As a result, the resulting policy is robust but tends to be overly conservative. For example, in Figs. 6–8, *full randomization* exhibits weak suppression of overshoot immediately after the driving condition is switched at 2 seconds. This behavior stems from insufficient learning of how to handle the backlash nonlinearity. When multiple uncertainties are addressed simultaneously during training, the learning for each individual uncertainty can become incomplete. Consequently, the comparison between the red and magenta lines supports the necessity of CL in which environmental uncertainties are progressively expanded and diversified.

The performance achieved by *Only MBC* exhibits significant variability depending on the uncertainty conditions. For instance, while performance comparable to the proposed method is attained in Fig. 6, a noticeable degradation—approaching the stability limit—is observed in Fig. 7. From the presence of the sim-to-real gap, such variability in performance is reasonable. Compared with the black line, the red line demonstrates the improved generalization capability of DRL through domain randomization.

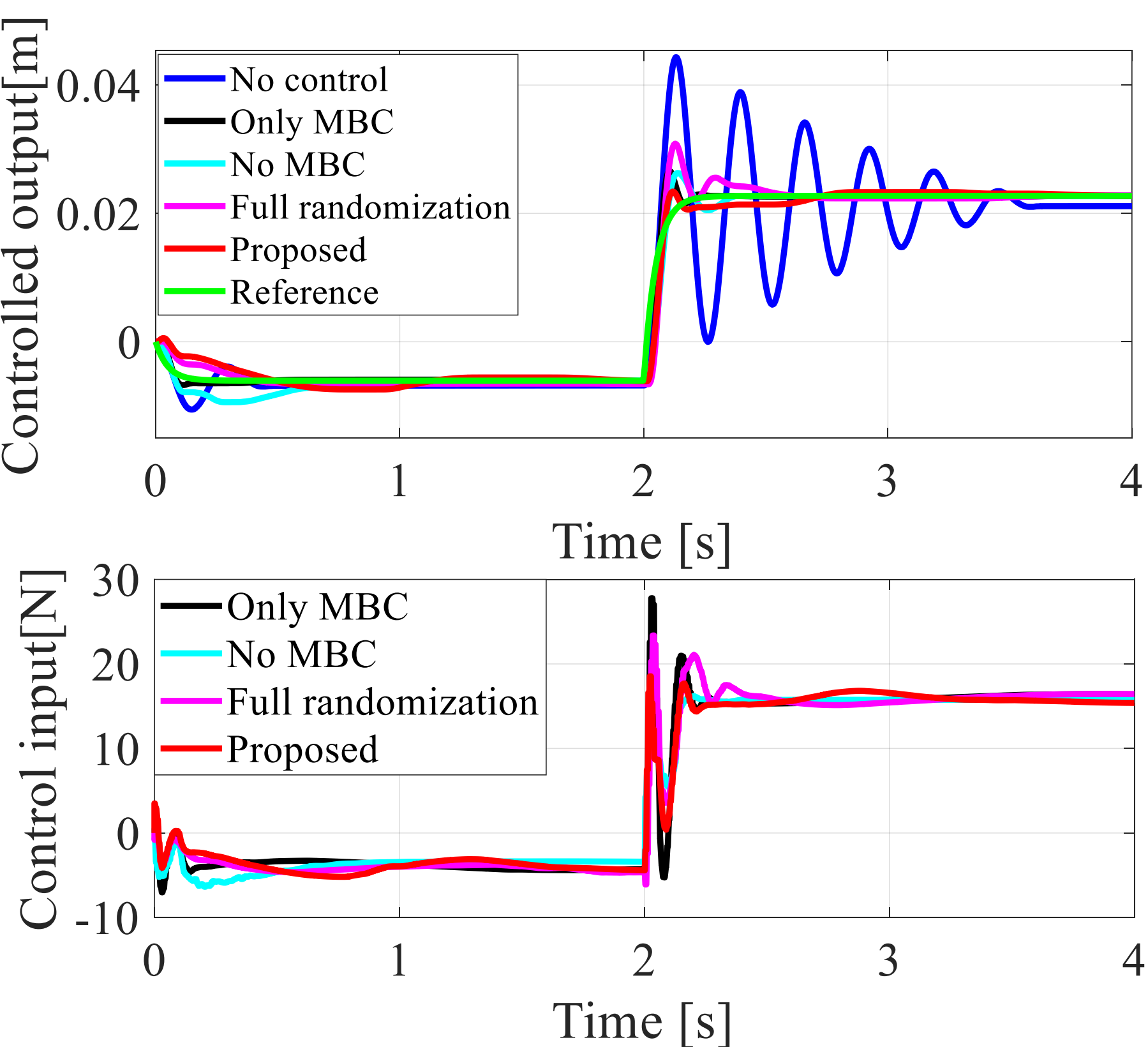

Fig. 5 Time responses of the vehicle body vibration (upper graph) and the control input (lower graph) for the linear nominal system.

Table 4 2-norm of the tracking error computed for the control results in Fig. 5.

| | 2-norm |
|---|---|
| Proposed method | 0.7174 |
| No MBC | 0.7517 |
| Full randomization | 0.8808 |
| Only MBC | 0.5890 |
| No control | 2.6940 |

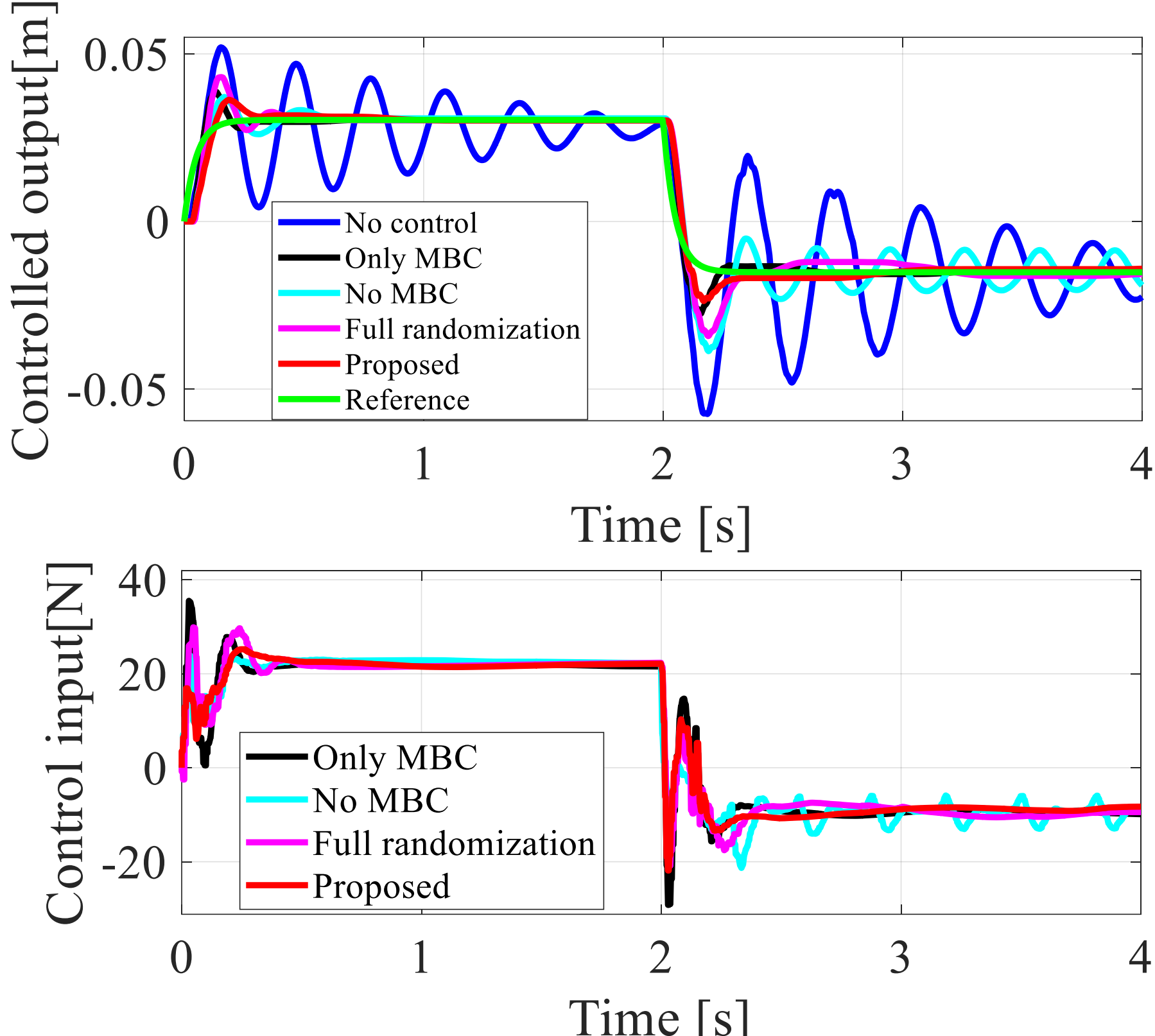

Fig. 6 Time responses of the vehicle body vibration (upper graph) and the control input (lower graph) with $M_B = M_B^{max}$, $M_E = M_E^{max}$, $C_G = C_G^{max}$, $C_D = C_D^{min}$, $C_C = C_C^{max}$, $y_{0-2s}^{r} = y_{0-2s}^{r\,max}$, $y_{2-4s}^{r} = y_{2-4s}^{r\,min}$, and $\delta = \delta_{min}$.

Table 5 2-norm of the tracking error computed for the control results in Fig. 6.

| | 2-norm |
|---|---|
| Proposed method | 1.4831 |
| No MBC | 2.5507 |
| Full randomization | 2.0811 |
| Only MBC | 1.4893 |
| No control | 6.4067 |

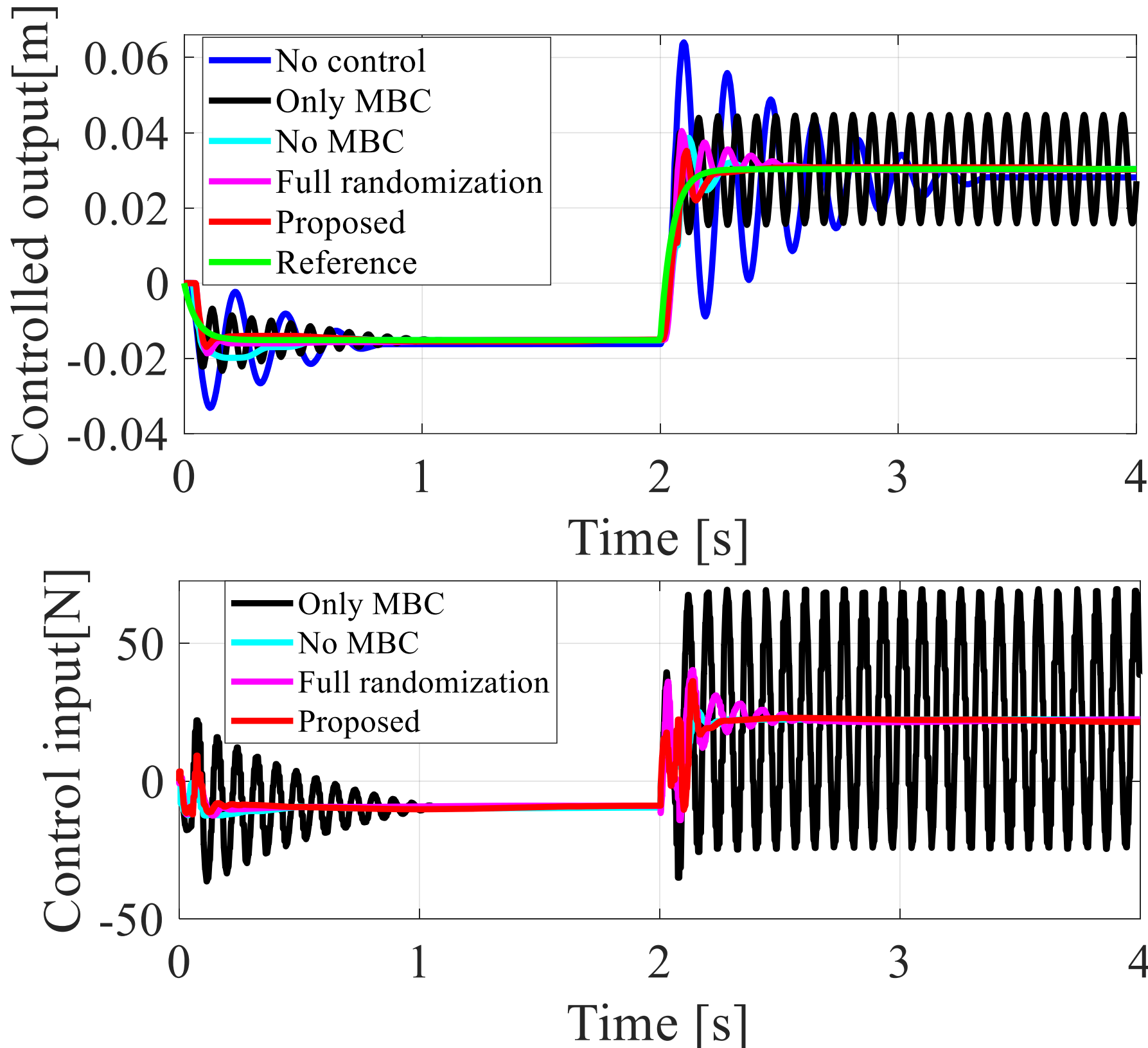

Fig. 7 Time responses of the vehicle body vibration (upper graph) and the control input (lower graph) with $M_B = M_B^{min}$, $M_E = M_E^{min}$, $C_G = C_G^{min}$, $C_D = C_D^{max}$, $C_C = C_C^{min}$, $y_{0-2s}^r = y_{0-2s}^{r\,min}$, $y_{2-4s}^r = y_{2-4s}^{r\,max}$, and $\delta = \delta_{max}$.

Table 6 2-norm of the tracking error computed for the control results in Fig. 7.

| | 2-norm |
|---|---|
| Proposed method | 0.8483 |
| No MBC | 1.0961 |
| Full randomization | 1.1414 |
| Only MBC | 3.3456 |
| No control | 4.1196 |

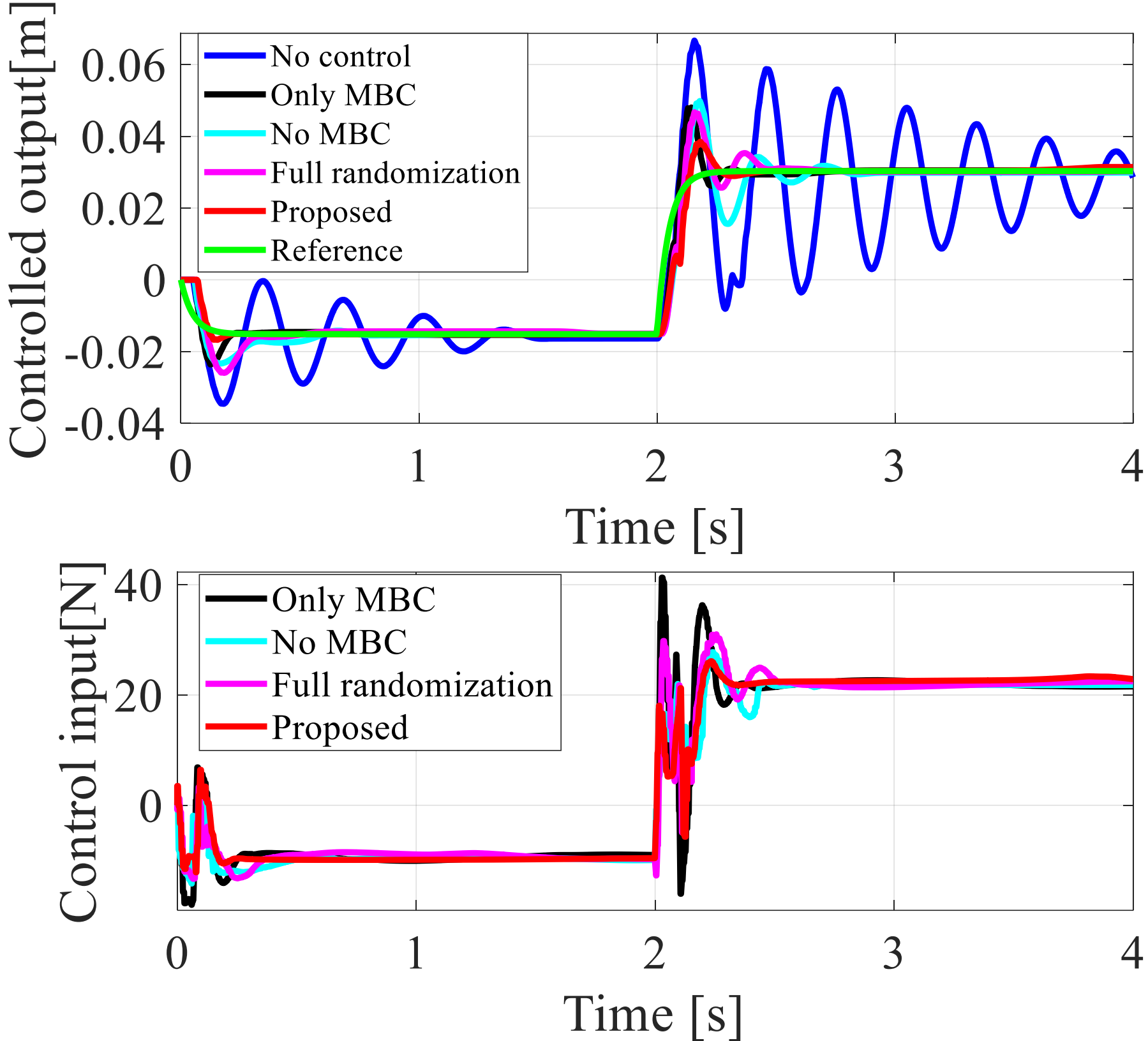

Fig. 8 Time responses of the vehicle body vibration (upper graph) and the control input (lower graph) with $M_B = M_B^{min}$, $M_E = M_E^{max}$, $C_G = C_G^{min}$, $C_D = C_D^{min}$, $C_C = C_C^{max}$, $y_{0-2s}^r = y_{0-2s}^{r\,min}$, $y_{2-4s}^r = y_{2-4s}^{r\,max}$, and $\delta = \delta_{max}$.

Table 7  2-norm of the tracking error computed for the control results in Fig. 8.

| | 2-norm |
|---|---|
| Proposed method | 1.3846 |
| No MBC | 2.0413 |
| Full randomization | 1.7692 |
| Only MBC | 1.4962 |
| No control | 5.8460 |

### 5.5. Monte Carlo simulation

To statistically evaluate the control performance of each method, Monte Carlo simulations were conducted. Specifically, each method was applied to 100 plant instances generated by randomly perturbing the uncertainty-related parameters, and their control performances were statistically analyzed. Figure 9 presents the mean time-history responses obtained over the 100 trials. In addition, Table 8 reports the mean and standard deviation of the 2-norm of the control error across the 100 simulations.

The proposed method achieves the best overall control performance, indicating the most stable control on average across various plant variations. In particular, as shown in Table 8, the proposed method attains the smallest standard deviation, demonstrating minimal variability in control performance with respect to plant variations. This is due to the curriculum that progressively and systematically acquires robustness against uncertainties, together with the stable policy updates enabled by EWC. By contrast, *Only MBC* exhibits a considerably larger standard deviation, indicating substantial performance variability and the existence of plant conditions under which the control performance deteriorates significantly. *Full randomization* achieves the second smallest standard deviation. Although jointly randomizing all the uncertainties often leads to a sub-optimal policy, it can still moderate the performance variability. The poor results of *No MBC* reflect insufficient learning maturity. As suggested in previous work [38], optimizing individual tasks without a foundational shared performance not only requires more training data but also increases vulnerability to abrupt task transitions.

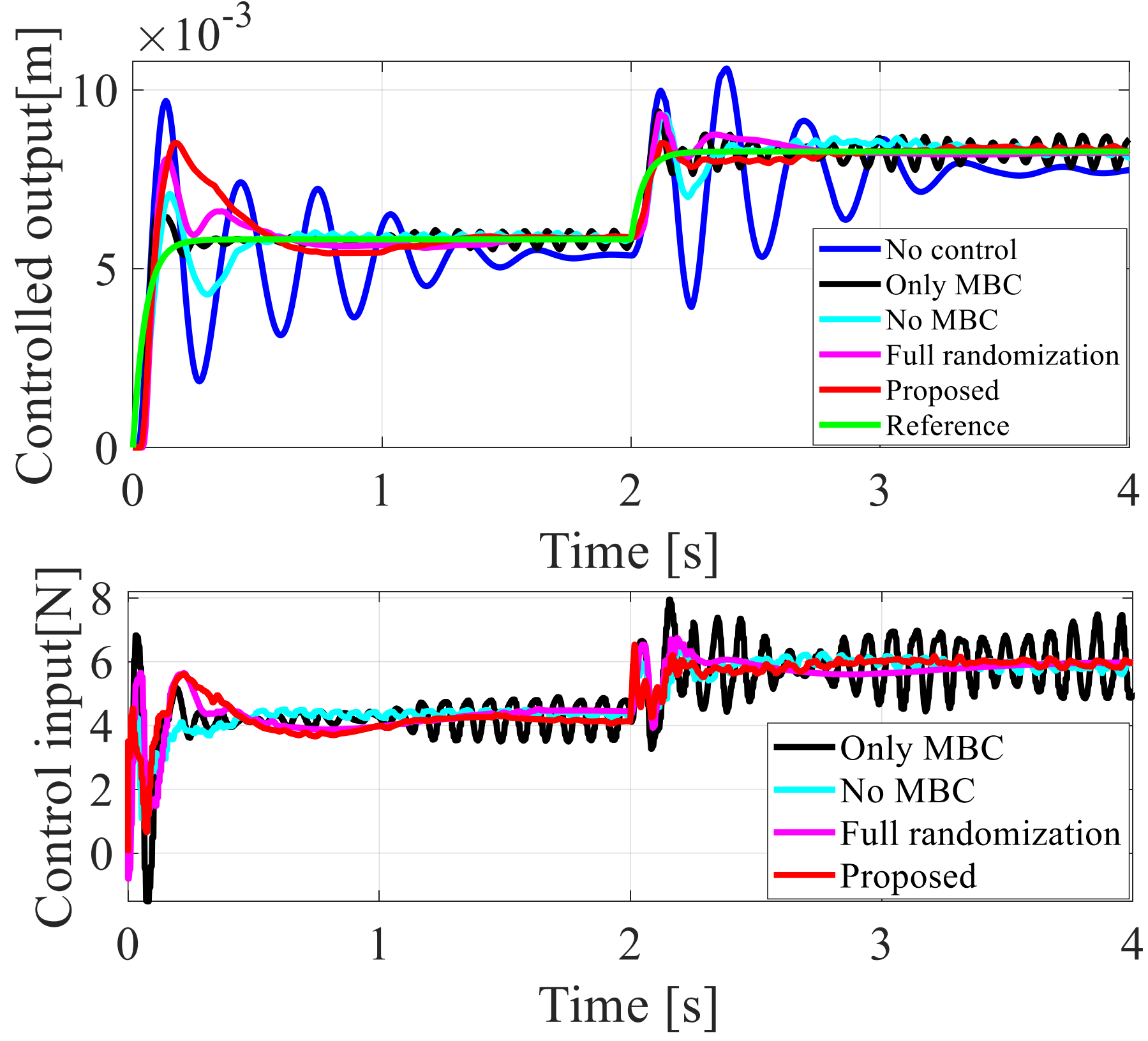

Fig. 9 Mean time responses of 100 trials when random variations are applied to the powertrain parameters.

Table 8 Statistics of 2-norm of 100 trials computed for the control results in Fig. 9.

| | 2-norm | |
|---|---|---|
| | Mean | Standard deviation |
| Proposed method | 0.6402 | 0.2201 |
| No MBC | 0.8085 | 0.3052 |
| Full randomization | 0.7364 | 0.3039 |
| Only MBC | 0.7153 | 0.7379 |
| No control | 1.9651 | 1.0938 |

## 6. Conclusion

This study proposed a curriculum-based continual uncertainty learning framework for dynamical systems with multiple intertwined uncertainties. The proposed approach decomposes the control problem into plant sets with progressively expanding uncertainties and formulates a continual learning problem to incrementally acquire robustness. To facilitate stable DRL training with reduced memory requirements, DDPG was integrated with online-EWC. In addition, the introduction of MBC was shown to ensure a shared baseline performance across the plant sets, improving the sample efficiency under highly diverse tasks. The practical application of the proposed scheme to an automotive powertrain control problem demonstrated superior robustness to multiple uncertainties and plant variations.
Future work will focus on experimental implementation on a real powertrain mechanism.

## Acknowledgment

This work was supported by the Japan Society for the Promotion of Science (JSPS) KAKENHI [Grant Number 23K13273].